# DreamKG: A KG-Augmented Conversational System for People Experiencing Homelessness

Javad M Alizadeh
*Temple University*
Philadelphia, USA
javad.m.alizadeh@temple.edu

Genhui Zheng
*University of Texas at Austin*
Austin, USA
genhuizheng@utexas.edu

Chiu C Tan
*Temple University*
Philadelphia, USA
cctan@temple.edu

Yuzhou Chen
*University of California Riverside*
Riverside, USA
yuzhou.chen@ucr.edu

Omar Martinez
*University of Central Florida*
Florida, USA
omar.martinez@ucf.edu

Philip McCallion
*Temple University*
Philadelphia, USA
philip.mccallion@temple.edu

Ying Ding
*University of Texas at Austin*
Austin, USA
ying.ding@austin.utexas.edu

Chenguang Yang
*University of California Riverside*
Riverside, USA
chenguang.yang@email.ucr.edu

AnneMarie Tomosky
*Temple University*
Philadelphia, USA
atomosky@temple.edu

Huanmei Wu
*Temple University*
Philadelphia, USA
huanmei.wu@temple.edu

**Abstract**

People experiencing homelessness (PEH) face substantial barriers to accessing timely, accurate information about community services. DreamKG addresses this through a knowledge graph-augmented conversational system that grounds responses in verified, up-to-date data about Philadelphia organizations, services, locations, and hours. Unlike standard large language models (LLMs) prone to hallucinations, DreamKG combines Neo4j knowledge graphs with structured query understanding to handle location-aware and time-sensitive queries reliably. The system performs spatial reasoning for distance-based recommendations and temporal filtering for operating hours. Preliminary evaluation shows 59% superiority over Google Search AI on relevant queries and 84% rejection of irrelevant queries. This demonstration highlights the potential of hybrid architectures that combines LLM flexibility with knowledge graph reliability to improve service accessibility for vulnerable populations effectively.

## I. INTRODUCTION

Access to timely, accurate information about essential community services, including shelters, food programs, healthcare, and social services is critical for PEH [1, 2]. However, significant information gaps persist, further compounded by unstable housing, limited digital access, and the transient nature of available resources [3, 4].

Existing solutions fall short in addressing these needs. General-purpose LLMs like ChatGPT, while conversationally fluent, suffer from hallucinations, confident but incorrect responses, with error rates exceeding 50% in clinical evaluations [5–7]. For vulnerable populations requiring reliable information about service locations, operating hours, and eligibility, such inaccuracies can have serious consequences. Current chatbot systems for homeless services rarely integrate structured knowledge or provide location-aware, time-sensitive recommendations [2, 8, 9]. While retrieval-augmented generation (RAG) and knowledge graph (KG) approaches improve factual accuracy [10–13], few systems combine these techniques to address the specific spatial-temporal information needs of PEH.

DreamKG addresses this gap through a KG-augmented conversational system designed to support community service navigation in Philadelphia. The system integrates Neo4j-based KGs, which encodes verified information on local organizations, with LLM-based query understanding to deliver reliable, explainable, location-aware

This work was supported by the NSF TIP, Grant No. 2333703.

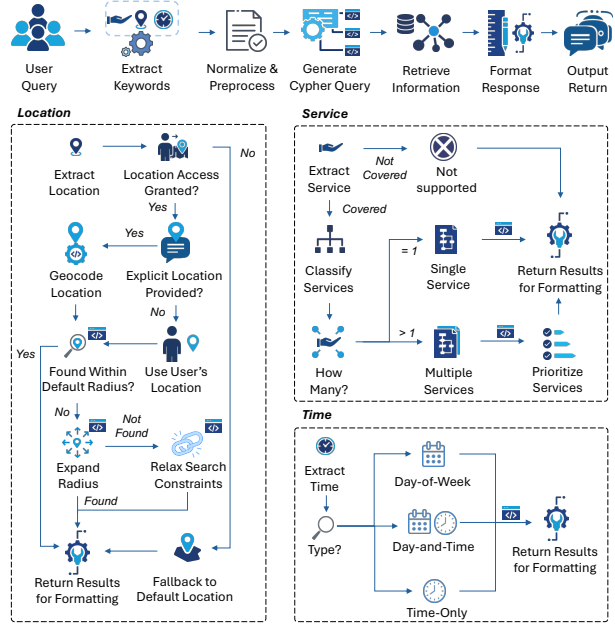

Fig. 1. System Architecture

recommendations. It incorporates spatial and temporal reasoning to account for distance preference and operating hours, while preserving conversational flexibility across multi-turn dialogues. This work advances KG–augmented AI by demonstrating how the integration of community-grounded data with spatial-temporal reasoning can improve service navigation for structurally vulnerable populations.

## II. DEMO DESCRIPTION

### A. System Overall Architecture

DreamKG employs a hybrid architecture (Fig. 1) that combines LLM-based natural language understanding with Neo4j KG retrieval. User queries are processed through six stages: (1) keyword extraction to identify service types, spatial cues, temporal constraints, and conversational context; (2) normalization and preprocessing, including geocoding locations and standardizing text; (3) Cypher query generation, which constructs schema-constrained database queries with fallback handling for inputs that cannot be mapped to the schema; (4) information retrieval through execution of queries against the KG; (5) response formatting, where results are structured based on query type; and (6) output rendering, which presents interactive visualizations alongside detailed service information.

### B. Data Source and Ontology

DreamKG integrates heterogeneous data from authoritative sources including FindHelp, OpenDataPhilly, the Office of Homeless Services (OHS), Department of Behavioral Health and Intellectual disAbility Services (DBHIDS), the Free Library of Philadelphia (FLP), and Social Security Administration (SSA). The Neo4j KG stores organization nodes (e.g., name, address, phone, services), location nodes (e.g., coordinates, ZIP codes, neighborhoods), service nodes (e.g., type, cost, eligibility), and temporal nodes (e.g., operating hours by day). A five-domain ontology, covering mental health services, social security offices, libraries, shelters, and food services, provides a structured framework for query interpretation.

### C. User Interaction and Key Features

Users interact with DreamKG through a Streamlit web interface (Fig. 2). The system supports conversational queries such as "Where can I get food near City Hall?" or "Are there any shelters open after 8pm on weekends?" Key functionalities include: (1) **Location-aware search**: Users can specify locations using addresses, landmarks (e.g., "City Hall"), or proximity terms (e.g., "near me," "within walking distance"). A spatial intelligence module geocodes inputs and ranks results by distance. (2) **Temporal filtering**: Queries can constrain results by operating hours and days (e.g., "open now," "weekends," "after 6pm"), with the system traversing temporal relationships in the KG to return only currently available services. (3) **Multi-turn conversation memory**: Follow-up queries such as "What about libraries?" or "Show me the closest one" leverage prior context without requiring users to restate preferences. (4) **Interactive map visualization**: Spatial queries are displayed on embedded Google Maps with distance indicators, addresses, and operating hours. Users can adjust starting points and view turn-by-turn directions. (5) **Structured service cards**: Each result displays expandable cards showing organization details, hours by day of week, service offerings, cost information, and contact details (Fig. 3). (6) **Transparent logging**: All queries, system decisions, and performance metrics are logged and available for download, supporting transparency and supporting continuous evaluation. Additional system resources (e.g., source code repository, a video walkthrough), are provided in the Appendix.

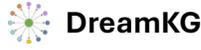

Fig. 2. User Interface

Fig. 3. Short Specific Response

## III. EVALUATION

We evaluated DreamKG against Google AI Mode using a structured, blind pairwise comparison judged by GPT-4.1-mini across 300 benchmark queries, including 200 domain-relevant and 100 out-of-scope queries.

**Benchmark Design.** The 200 relevant queries were synthetically generated to reflect realistic helper-oriented search behavior across five service categories: food banks, mental health services, shelters, public libraries, and Social Security offices. Queries were constructed under a full factorial design, varying with two types of temporal expressions (implicit or explicit clock times), four location specifications (ZIP code, neighborhood name, street address, and ambiguous environmental cues), and five language styles (interrogative, declarative, search-style, causal/reflective, and community-oriented). Relevancy was defined as alignment with Philadelphia-based homeless service navigation. The 100 irrelevant queries consisted of general-purpose questions (*e.g.*, cooking, technology support, travel) unrelated to social services.

**Baseline.** Google AI Mode's generated summary response served as the baseline. DreamKG responses combined a concise direct answer with a longer, more detailed contextual explanation, evaluated together as a single output.

**Judge Setup.** For each relevant query, the judge received anonymized A/B responses and selected the response that better supports a real-world helper seeking clear, actionable, and high-confidence service information. Evaluation criteria were prioritized as follows: (1) location specificity (proximity to the requested ZIP code); (2) service relevance (*e.g.*, domain-specific providers rather than generic resources); (3) operational detail (availability of address, phone, hours, and service descriptions); (4) structural clarity (information organization and readability); (5) actionability (ability to directly contact or visit a provider); (6) signal-to-noise ratio (focus and conciseness); and (7) overall accuracy and plausibility (realistic data).

**Example Cases.** For "*Where can I refer someone to a food bank in 19104 earlier today?*", DreamKG returned a specific organization with complete operational details, including address, phone number, and hours. In contrast, for "*Is there a walk-in mental health crisis center in 19124 for someone I'm trying to help earlier today?*", Google AI Mode was preferred because it explicitly confirmed walk-in availability and same-day hours. This example highlights a key limitation of DreamKG: its reliance on scheduled operating hours limits its ability to capture unplanned real-time changes such as temporary closures or disruptions.

## IV. NOVELTY

DreamKG introduces three key innovations for community services navigation among PEH: (1) *compositional query construction*, which jointly models service type, spatial proximity, and temporal constraints into multi-turn interactions, which can decompose complex queries with conversational reference resolution; (2) *constrained KG retrieval* that combines geocoding, coordinate-based graph traversal, and time-window filtering, with fallback handling across heterogeneous data sources; (3) *community-engaged data curation,* supported by a modular system architecture that allows the iterative integration of heterogeneous service data provided by community partners.

## V. CONCLUSION

DreamKG demonstrates the potential of KG-augmented conversational AI to reliably support vulnerable populations requiring accurate, location-aware, and time-sensitive information. In preliminary evaluation using an independent LLM-based judge, DreamKG outperformed Google Search AI across six criteria: location specificity, service relevance, operational detail, structural clarity, actionability, and signal-to-noise ratio, showing 59% superiority on relevant queries and 84% appropriate rejection of irrelevant queries. The system was further evaluated through Community Advisory Board (CAB) review, aligning with established best practices [14–18] and ensuring responsiveness to real-world service navigation challenges while mitigating risks of misrepresentation or harm. The work highlights how structured knowledge representations enable precise spatial-temporal reasoning, how conversational memory supports multi-turn dialogues, and how transparent query construction fosters trust with populations facing systemic barriers. Future work will involve multi-site evaluations across diverse urban settings to assess the system's impact on service access, care coordination, and health and social outcomes among people experiencing homelessness.

## Appendix

### A. System Access and Resources

DREAM KG is publicly accessible and fully open-source to support replication, extension, and community-engaged deployment:

- Live Demonstration:
  https://dream-kg.streamlit.app
- Source Code Repository:
  https://github.com/JavadMAlizadeh/DreamKG
- Video Walkthrough:
  https://www.youtube.com/watch?v=S-pOfgtU4ik
- Proto-OKN Project Context:
  https://www.proto-okn.net/

### B. System Interface

The graphical user interface offers multiple interaction layers for navigation and service discovery. Fig. 4 demonstrates the system handling queries that span multiple service categories simultaneously. Fig. 5 displays additional organization details such as contact info and hours. Fig. 6 shows transport mode selection for turn-by-turn directions. Fig. 7 provides an embedded street view for previewing a location.

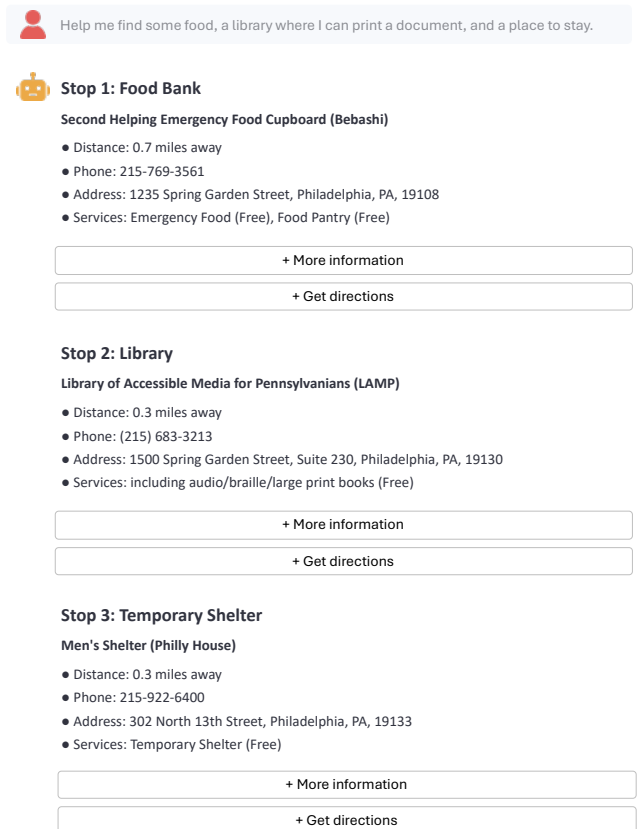

Fig. 4. Sample response for queries requesting multiple services

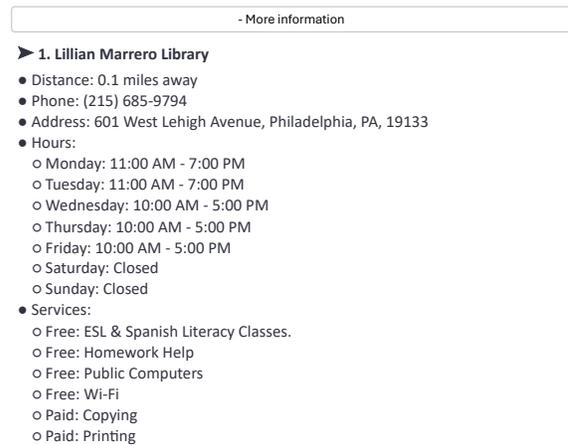

Fig. 5. More information

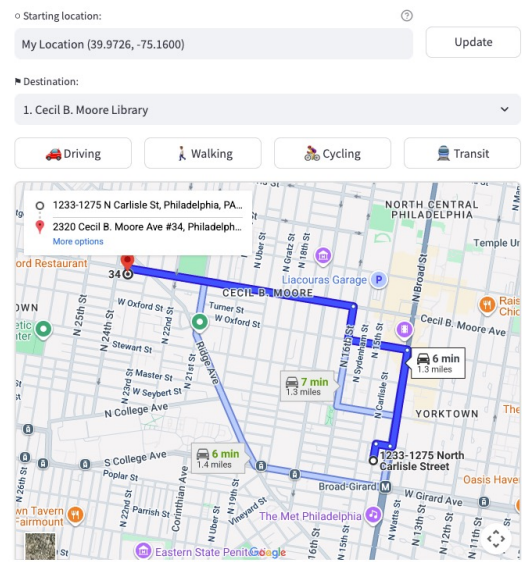

Fig. 6. Navigation information

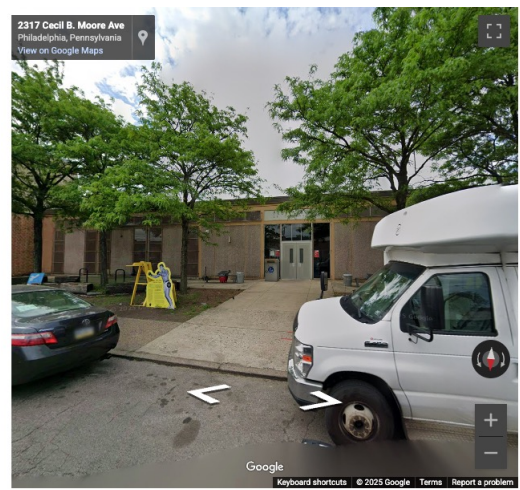

Fig. 7. Street view